\documentclass[10pt,journal,compsoc]{IEEEtran}

\ifCLASSOPTIONcompsoc
  \usepackage[nocompress]{cite}
\else
  \usepackage{cite}
\fi
\usepackage{graphicx}
\usepackage{amssymb}
\usepackage{amsmath}
\usepackage{booktabs}
\usepackage{multirow}
\usepackage{txfonts}
\usepackage{mathrsfs}

\usepackage{amsthm}
\usepackage{color}
\usepackage{xcolor}
\ifCLASSINFOpdf
\else
\fi
\begin{document}
%
\title{PointSmile: Point Self-supervised Learning via Curriculum Mutual Information}
%
%
%
%

\author{Xin~Li,
        Mingqiang~Wei,
        and Songcan~Chen
\IEEEcompsocitemizethanks{\IEEEcompsocthanksitem X. Li, M. Wei and S. Chen are with the School of Computer Science and Technology, Nanjing University of Aeronautics and Astronautics, and also with the MIIT Key Laboratory of Pattern Analysis and Machine Intelligence, Nanjing, China (e-mail: thea\_lee@nuaa.edu.cn; mingqiang.wei@gmail.com; s.chen@nuaa.edu.cn). 
}
}

\IEEEtitleabstractindextext{%
\begin{abstract}
Self-supervised learning is attracting wide attention in point cloud processing. However, it is still not well-solved to gain discriminative and transferable features of point clouds for efficient training on downstream tasks, due to their natural  sparsity and irregularity. 
We propose \textit{PointSmile}, 
a reconstruction-free \textbf{s}elf-supervised \textbf{le}arning paradigm by maximizing curriculum \textbf{m}utual
\textbf{i}nformation (CMI) across the replicas of \textbf{point} cloud objects.
From the perspective of \textit{how-and-what-to-learn}, PointSmile is designed to imitate human curriculum learning, i.e., starting with an easy curriculum and gradually increasing the difficulty of that curriculum. 
To solve  ``how-to-learn", we introduce curriculum data augmentation (CDA) of point clouds. CDA encourages PointSmile to learn from easy samples to hard ones, such that the latent space can be dynamically affected to create better embeddings.
To solve ``what-to-learn", we propose to maximize both feature- and class-wise CMI, for better extracting discriminative features of point clouds. 
Unlike most of existing methods, PointSmile does not require a pretext task, nor does it require
cross-modal data to yield rich latent representations. We demonstrate the effectiveness and robustness of PointSmile in downstream tasks including object classification and segmentation. Extensive results show that our PointSmile outperforms existing self-supervised methods, and compares favorably with popular fully-supervised methods on various standard architectures.

\end{abstract}

\begin{IEEEkeywords}
PointSmile, Self-supervised learning, Curriculum mutual information.
\end{IEEEkeywords}}

\maketitle

\IEEEdisplaynontitleabstractindextext

%
\IEEEpeerreviewmaketitle

\section{Introduction}
\label{sec:intro}

There is an increasing demand to capture the real world by 3D sensing techniques for applications such as Metaverse and digital twins \cite{zhuzhetvcg}.
The captured scenes are often represented in a simple and flexible form, i.e., point cloud \cite{Wei2022AGConv}.
Recent years have witnessed considerable efforts of using deep learning to understand point clouds \cite{gulipeng}.
The first step for point cloud understanding is to extract discriminative geometric features \cite{xianzhili}, which is referred to as geometric representation learning (GRL).
Ideally, when fed with sufficient annotated data, GRL will become powerful that can combine various neural networks, e.g., PointNet \cite{DBLP:conf/cvpr/QiSMG17}, PointNet++ \cite{DBLP:conf/nips/QiYSG17}, and DGCNN \cite{DBLP:journals/tog/WangSLSBS19}, to facilitate the downstream tasks such as classification and segmentation.
However, real-world scenarios often lack labeled 3D scans, and human annotations of those scans are very laborious due to their irregular structures \cite{honghuachen}. Although training on synthetic scans is promising to alleviate the shortage of labeled real-world data, such trained GRL will inevitably suffer from domain shifts.   


Self-supervised learning, as an unsupervised learning paradigm, can relieve the shortcomings of supervised models, and is validated in 2D fields \cite{DBLP:conf/icml/ChenK0H20, DBLP:conf/nips/GrillSATRBDPGAP20,DBLP:journals/corr/abs-2003-04297}. 
This motivates the recent surge of interests in extracting powerful features by self-supervised learning \cite{DBLP:conf/cvpr/EckartYLK21,DBLP:journals/corr/abs-2202-04241,DBLP:journals/corr/abs-2203-14084} for 3D point clouds \cite{DBLP:conf/cvpr/YangFST18, DBLP:conf/iccv/HanWLZ19, DBLP:conf/icml/AchlioptasDMG18, DBLP:conf/aaai/HanSLZ19}.
Most of the existing self-supervised learning methods follow the widely-used encoder-decoder architecture, in which the update of their encoders' parameters depends on the reconstruction of point cloud objects in the decoder. However, i) reconstructing those 3D objects is not always attainable, due to the discrete nature of point clouds; ii) the unimodal losses of such as mean squared error and cross-entropy are not feasible to recover various geometric details in the original data; 
and iii) these models are computationally intensive to formulate the complex relationships in the data and tough to optimize.
\begin{figure}[t]
  \centering
  
   \includegraphics[width=1.0\linewidth]{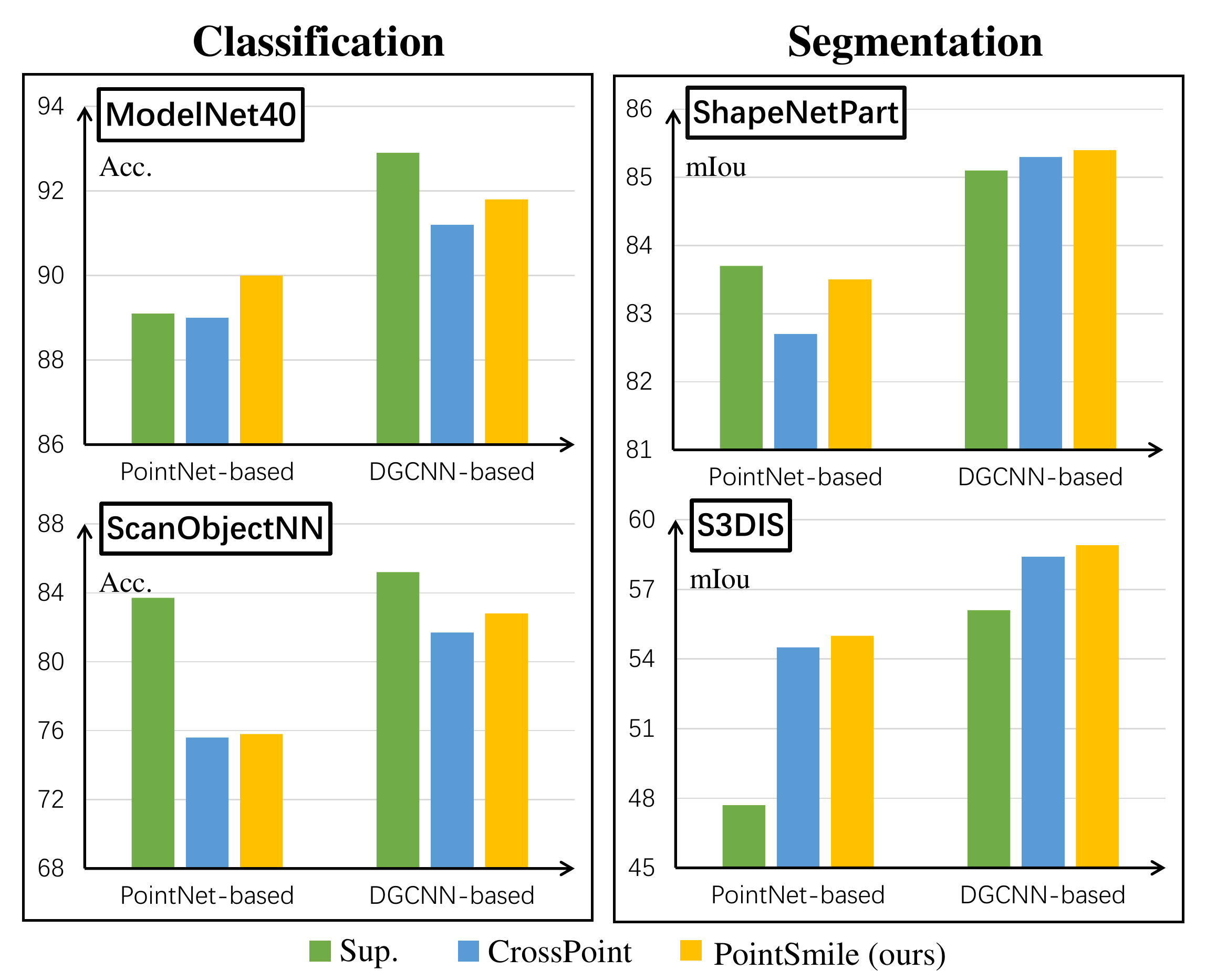}

   \caption{Self-supervised learning is challenging for point clouds.  Benefiting from curriculum mutual information, our single-modal PointSmile illustrates clear improvements over the cross-modal CrossPoint \cite{DBLP:conf/cvpr/AfhamDDDTR22}, and is comparable to the supervised methods \cite{DBLP:conf/cvpr/QiSMG17}, \cite{DBLP:journals/tog/WangSLSBS19} for various downstream tasks. Left: the accuracy of linear SVM classification on ModelNet40 \cite{DBLP:conf/cvpr/WuSKYZTX15} and ScanObjectNN \cite{DBLP:conf/iccv/UyPHNY19}. Right: the mean IoU of part segmentation (top) and semantic segmentation (bottom) on ShapeNetPart \cite{DBLP:journals/tog/YiKCSYSLHSG16} and S3DIS \cite{armeni20163d}. Two backbone networks PointNet \cite{DBLP:conf/cvpr/QiSMG17} and DGCNN \cite{DBLP:journals/tog/WangSLSBS19} are utilized.
   }
   \label{fig:overview}
\end{figure}

Imagine how our teachers taught us complex knowledge when we were freshmen. They may make a plan of curriculum learning, in which the easy and intuitive knowledge will be first presented, followed by the hard and abstract knowledge \cite{avrahami1997teaching}. Such a curriculum makes the students leverage previously learned knowledge to easily learn the increasingly difficult contents, thereby reducing the abstraction of new knowledge. 
%
We attempt to absorb such wisdom of curriculum learning for extracting discriminative features of unlabelled point clouds that are transferable to downstream tasks. 
To imitate the curriculum learning of humans for GRL, we need to know how to learn and what to learn.

For \textit{how-to-learn}, we introduce curriculum data augmentation (CDA) to construct two types of the replicas for each unlabeled 3D object, namely easy samples and hard samples. 
Thus, the geometric representation can learn from these easy samples to the hard ones gradually, such that the latent space is dynamically affected to create better embeddings.
For \textit{what-to-learn}, we propose to learn the maximization of curriculum mutual information (CMI) across the replicas
of an unlabeled 3D object, which encourages to better extract discriminative features of point clouds. Different with the original mutual information, CMI is maximized jointly from two aspects, i.e., feature- and class-wise CMI. First, we maximize feature-wise CMI to make the features belonging to the same class more similar and dense in the feature space.  Second, maximizing only feature-wise CMI may lead to different classes getting further and further apart in the feature space. Thus, we formulate class-wise CMI to emulate the human perception that two similar objects but with different labels can have closer semantic features (e.g., a single sofa and a chair with a backrest).
Herein, the term \textit{class} means the representation cluster divided in an unsupervised manner while it is not the real classification label.
By maximizing feature- and class-wise CMI jointly, the inner- and intra-class representations can be distributed uniformly in the feature space. 

We propose a reconstruction-free, self-supervised geometric representation learning paradigm via maximizing CMI, dubbed PointSmile. Instead of using extracted representations to reconstruct 3D objects, we enlarge the correlation between the representations obtained from different replicas of the same object. PointSmile is conceptually simple, easy to implement, and it learns useful geometric representations. Compared to existing methods, it does not require a complex pretext task, nor does it require cross-modal data. With even just one linear layer, it provides classification results comparable to the supervised methods.

We evaluate our approach on multiple downstream tasks. First, we perform shape classification in ModelNet40 \cite{DBLP:conf/cvpr/WuSKYZTX15}, a synthetic object dataset. Second, we perform shape classification in ScanObjectNN \cite{DBLP:conf/iccv/UyPHNY19}, a real-world object dataset, to assess its transferability. Third, part segmentation and semantic segmentation are performed to verify the ability of our PointSmile to capture essential fine-grained features.
In our experiments, we employ two widely used point cloud networks as our feature extractors, to assess the generality of our approach. As shown in Figure~\ref{fig:overview}, our approach (yellow) outperforms CrossPoint \cite{DBLP:conf/cvpr/AfhamDDDTR22} (blue) for all downstream tasks in different datasets, and it can even outperform  the supervised methods (green).

The main contributions of our work are three-fold:
\begin{itemize}
    \item
    We propose PointSmile, a new self-supervised learning paradigm via curriculum mutual information maximization. PointSmile possesses higher abstraction and keeps the invariance of geometric transformations. PointSmile is decoder-free, which avoids the complicated and unstable reconstruction of 3D objects, and can be flexibly combined with mainstream neural networks, such as PointNet/PointNet++, and DGCNN.
    \item We propose a ``how-and-what-to-learn" strategy to i) upgrade the degree of difficulty in data augmentation step by step called curriculum data augmentation, and ii) to maximize the curriculum mutual information. Such a strategy ensures to effectively learn the discriminative features of point clouds without any annotation.
    
    \item 
    We demonstrate the PointSmile's efficacy through extensive evaluations in several downstream tasks, i.e., object classification and part segmentation. It not only achieves a better performance than its competitors but also demonstrates a better generalization capability. What is more, we analyze the superiority of our approach by comparing it to existing self-supervised learning methods.
\end{itemize}

\section{Related Work}
Representation learning (RL), aiming to automatically discover the feature patterns in the data, is a very important aspect of deep learning. 
Although achieving great success in image processing,
RL is still not well explored for processing point clouds, which is referred to as geometric representation learning (GRL).
We will review current supervised and self-supervised GRL methods, followed by mutual information maximization.
\begin{figure*}[]
    \centering
    \includegraphics[width=1\textwidth]{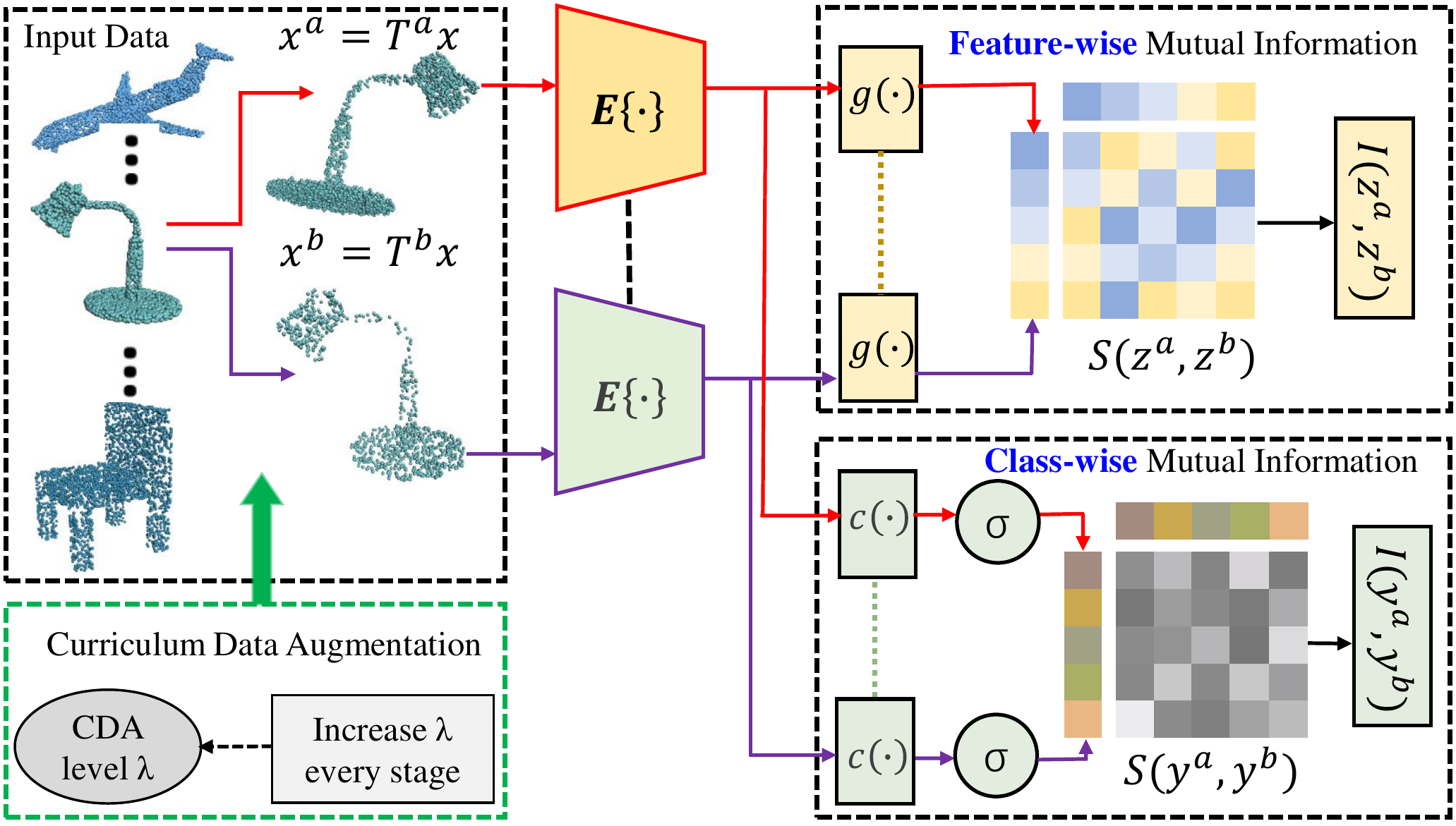}
    \caption{Overview of PointSmile. PointSmile is designed to imitate how humans to learn professional knowledge via curriculum learning. It consists of three main components, i.e., i) curriculum data augmentation to construct easy and hard replicas of each 3D object, and to increasingly add the portion of hard replicas during learning; ii) a shared encoder $E$ to learn geometric representations, and iii) two curriculum mutual information (CMI) modules to maximize the feature-wise and class-wise CMI jointly. $x$ denotes an input point cloud batch and $x^{a}$, $x^{b}$ denote two different replicas of $x$ obtained from CDA.}
    \label{fig:full_model}
\end{figure*}

\subsection{Supervised Representation Learning}
Many supervised GRL methods have emerged, and we divide them into conversion-based,
point-based and graph-based methods.

\textbf{Conversion-based methods} convert point clouds into regular 2D grids \cite{DBLP:conf/cvpr/ZhouCF000WW020}, 3D voxels \cite{maturana2015voxnet,DBLP:conf/cvpr/WuSKYZTX15} or develop hand-crafted feature descriptors \cite{DBLP:journals/tvcg/ChenWSXW20,DBLP:journals/cad/LiZFXWWH20}, in which traditional 2D/3D CNNs can be smoothly operated. 
KD-Net \cite{klokov2017escape} uses more efficient data structures and skip the calculation of empty voxels. PointGrid \cite{le2018pointgrid} integrates point and mesh representations by sampling a constant number of points in each embedded volumetric grid cell, thereby efficiently extracting geometric details by 3D CNNs. 
MVCNN \cite{su2015multi} and RangeNet++ \cite{milioto2019rangenet++} generate multi-view features by rendering point clouds as 2D images for different downstream tasks.
However, these conversion-based techniques are sensitive to noise and outliers, and are hard to capture fine-grained geometric details. Also, they often introduce excessive memory cost.

\textbf{Graph-based methods} regard points as the nodes of a graph and create edges based on their spatial/feature relationships \cite{kipf2016semi}. 
KCNet \cite{shen2018mining} defines kernels based on Euclidean distances and geometric affinities of neighboring points.  DGCNN \cite{DBLP:journals/tog/WangSLSBS19} gathers the nearest neighboring points in the feature space, and uses EdgeConv to dynamically identify semantic cues for feature extraction. 
3D-GCN \cite{lin2020convolution} develops deformable kernels, extracting local 3D features across scales and focusing on shift and scale-invariant properties in point cloud analysis. AdaptConv \cite{DBLP:conf/iccv/ZhouFFW0L21} exploits adaptive kernels to replace the weight-sharing operation used in standard graph convolution, and adaptively establishes diverse connections between different points in the local neighborhood from different semantic parts. 

\textbf{Point-based methods} handle the irregularity of point
clouds by directly
manipulating them, rather than introducing various intermediate representations.
As the pioneer, PointNet leverages multi-layer perceptrons independently on each point to directly process points. 
During pooling operations, PointNet and DeepSets \cite{DBLP:conf/nips/ZaheerKRPSS17} abandon considerable local features  that are indispensable to describing 3D shapes. PointNet++ \cite{DBLP:conf/nips/QiYSG17} addresses the weakness of PointNet by exploiting a hierarchical structure to extract local features and introducing sampling and grouping operations. 
Similar ideas exist in PointCNN \cite{DBLP:conf/nips/LiBSWDC18}, and PointConv \cite{DBLP:conf/cvpr/WuQL19}. These methods first establish the topological relationship between points to extract local semantic features, and then aggregate the features by concatenating the features or improving the representation capability with RNN. PointMLP \cite{ma2022rethinking} supposes that a sophisticated local geometric extractor may not be that important for performance and only uses residual feed-forward MLPs, without any other local feature exploration. 
Transformers, in encoder and/or decoder configurations, have been successfully applied in NLP and CV. Many efforts also use Transformers for point cloud processing, such as Point Transformer \cite{DBLP:conf/iccv/ZhaoJJTK21} and PCT \cite{guo2021pct}.

Generally, for the supervised methods, there is an urgent need to solve the problem of how to obtain accurate labels efficiently.

\subsection{Self-supervised Representation Learning}
\textbf{Generative methods} learn features via self-reconstruction, which first encodes the point cloud into a feature or distribution and then decodes it back to the point cloud \cite{DBLP:conf/cvpr/YangFST18,YonghengZhao20183DPC,DBLP:conf/iccv/HanWLZ19,DBLP:conf/icml/AchlioptasDMG18,DBLP:conf/iccv/HassaniH19}. Recently, a wide variety of self-supervised methods are proposed based with Transformer. For example, Point-Bert \cite{DBLP:conf/cvpr/YuTR00L22} predicts discrete tokens and Point-MAE \cite{DBLP:journals/corr/abs-2203-06604} randomly masks patches of the input point clouds and reconstructs the missing points. An alternative to generative methods is to use generative adversarial networks for generative modeling \cite{han2019view, achlioptas2018learning}. 
However, one disadvantage of these methods is that they require reconstruction or generation of 3D shapes. As mentioned earlier, reconstructing the shape may be expensive or impossible.

\textbf{Discriminative methods} learn point cloud representations based on auxiliary hand-crafted prediction tasks. For this class of methods, it is not the optimal selection to reconstruct the 3D shapes directly from the representation.
Jigsaw3D \cite{DBLP:conf/nips/SauderS19} uses a 3D Jigsaw puzzle as the self-supervised learning task and trains an encoder for downstream tasks through contrastive techniques.
PointContrast \cite{DBLP:conf/cvpr/XieLCT18} proposes a pretext task, in which the representation of a single point cloud from different views should remain consistent and focus on high-level scene understanding tasks. Based on this task, it investigates a unified comparative paradigm framework for 3D representation learning. CrossPoint \cite{DBLP:conf/cvpr/AfhamDDDTR22} combines information from both 3D and 2D modalities, focusing on powerful features shared between the different modalities. It demands difficult-to-obtain point cloud rendering outcomes, although it is straightforward and efficient. To make contrastive learning tasks easier, Du et al. \cite{DBLP:conf/mm/DuGHL21} use self-similar point cloud patches from a single point cloud as positive or negative examples and actively learn hard negative examples near positive samples for discriminative feature learning. STRL \cite{huang2021spatio} is a direct extension of BYOL \cite{grill2020bootstrap} to 3D point clouds, which learns representations through the interaction of online and target networks.

Besides, some methods attempt to combine discriminative methods with another self-supervised strategy: clustering. Zhang et al. \cite{DBLP:conf/3dim/ZhangZ19} combine contrastive method and clustering offline to learn representation gained a yields a promising result. However, it needs to be trained in two stages.

These discriminative approaches focus mainly on the association of positive and negative samples or on the design of pretext tasks. Therefore, the quality of these factors also influences the learned features. In contrast to the existing works which leverage generative and discriminative approaches, we introduce a more straightforward way of using mutual information that does not require a decoder and does not require redundant pretext tasks, but yields better representation.

\subsection{Mutual Information Maximization}

Information theory has a long history of being applied as a tool for training deep networks for 2D images. IMSAT \cite{DBLP:conf/icml/HuMTMS17} uses data augmentation to impose the invariance on discrete representations by maximizing mutual information between data and its representation. More recently, Oord et al. \cite{DBLP:journals/corr/abs-1807-03748} propose the framework of Contrastive Predicting Coding (CPC), which combines predicting future observations with a probabilistic contrastive loss. CPC uses the embeddings to capture maximal information about future samples. Deep InfoMAX \cite{DBLP:conf/iclr/HjelmFLGBTB19} maximizes the mutual information between input data and learned high-level representations and has the advantage of performing orderless autoregression. However, it computes mutual information over continuous random variables, which requires complex estimators. In contrast, IIC \cite{DBLP:conf/iccv/JiVH19} obtains mutual information of discrete variables with simple and exact computations. DGI \cite{velickovic2019deep} relies on maximizing mutual information between patch representations and corresponding high-level summaries of graphs. Our methods are inspired, but unlike these approaches, we extend them to 3D representations. We consider maximize mutual information not only from a local perspective, but also from a global perspective.

\section{Method} 
\label{sec:method}

\subsection{Overview}
Imagine when we start a plan of
curriculum learning, the easy and intuitive knowledge will be first taught, followed by the hard and abstract knowledge. Such wisdom
of curriculum learning inspires our GRL paradigm. That is, by proposing a ``how-and-what-to-learn" strategy, we design a self-supervised GRL framework, called PointSmile.

At the top level, we show in Figure~\ref{fig:full_model} our PointSmile, a decoder-free model that learns discriminative features from 3D point clouds in a self-supervised manner. 
PointSmile consists of three main components, i.e., i) curriculum data augmentation (CDA) to construct easy and hard replicas of each 3D object, and to increasingly add the portion of hard samples during learning; 
ii) a shared encoder $E$ to learn geometric representations, and iii) two CMI modules (i.e., feature-wise CMI and class-wise CMI) to maximize the mutual information between features extracted from independently-augmented pairs of each point cloud.


\subsection{Curriculum Augmented Pairs from CDA}

\begin{figure}[t]
  \centering
  
   \includegraphics[width=1.0\linewidth]{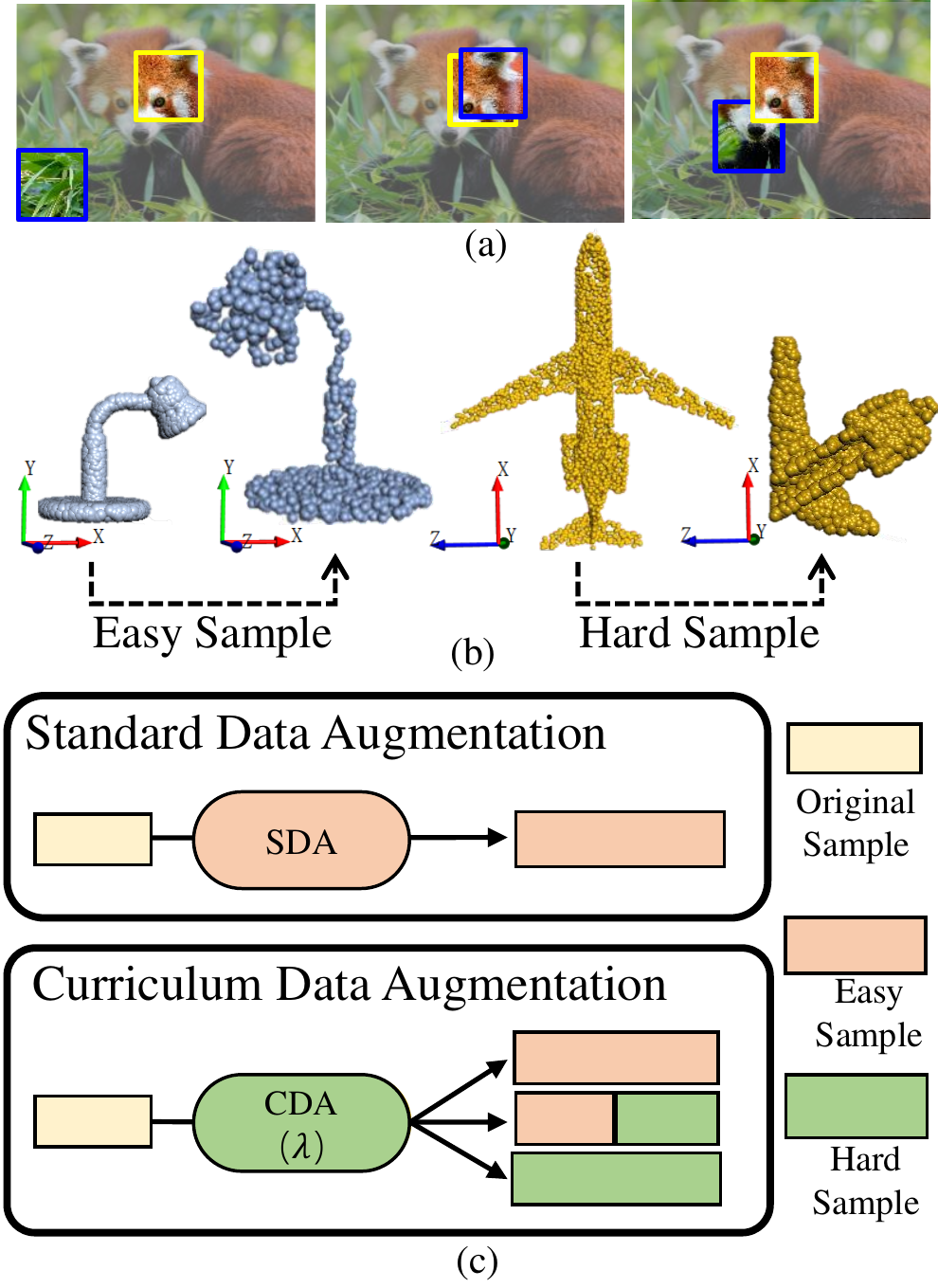}

   \caption{Comparison of different ways to obtain augmented samples.
   (a) An example comparison of three different regimes for the augmented sample pairs in an image. (b) Comparison of easy augmented sample (easy sample) and hard augmented sample (hard sample). We show the geometric transformation of the sample with a uniform xyz coordinate axis. (c) Comparison of SDA and CDA. The length of the rectangle represents the number of augmented samples with the corresponding color. Please note that $\lambda$ is an array that grows over the epochs. 
   }
   \label{fig:cda}
\end{figure}


\textbf{Motivation.}
We observe that a GRL model is prone to overfitting by purely given easy samples, since these easy samples only involve simple geometric transformations. Therefore, we can mine hard samples to enhance the GRL model. However, if feeding the GRL model with the combination of easy and hard samples directly, it might confuse the `learner' and lead to harms of model training at the initial stage. It is reasonable to learn from easy samples and increasingly add hard samples. 


\textbf{Curriculum Data Augmentation.}
We leverage Curriculum Data Augmentation (CDA) to update the difficulty of data augmentation increasingly. 
The principle of CDA is that the augmented data should resemble the original data. Under this principle, we control the CDA intensity, ensuring the augmented pairs to have lower CMI. This allows the model to gradually accumulate more sophisticated information during the learning process.

We construct two types of replicas for each point cloud.
The first type of replicas is called easy samples and are generated by standard data augmentation (SDA) operations including random scale, translation, down-sampling, jitter and rotation. As shown in the left part of Figure~\ref{fig:cda} (b), the easy sample contains geometric transformation that shares much information with the original object.
Thus, it is easy to figure out the two point clouds are from the same object. 
The second type that shares the less information is called hard samples and is generated by using the techniques including random X/Y axis flipping, shift, cuboid augmentation and drop cuboid.
For example, it is hard to identify the replica which remains only the middle-and-rear part by cutting the original part of an airplane as shown in the right part of Figure~\ref{fig:cda} (b). Easy samples make up easy sample pairs, and so do hard sample pairs.
These hard sample pairs have lower CMI than the easy sample pairs, and mainly preserve the downstream task-relevant high-level information. 

To clearly explain why the easy sample pair and the hard sample pair have different CMI, we first explain this phenomenon in 2D, and then change to 3D. We test two different samples (blue frames in Figure~\ref{fig:cda} (a)) as a sample pair of the same image. 
In the right image of Figure~\ref{fig:cda} (a), the sample pair belongs to different parts of the red panda but contain a few overlapped areas of the foreground. The pair is an optimal mix to provide the most independent information needed by the model. In the left image of Figure~\ref{fig:cda} (a), the sample pair is from the foreground and background, respectively. It shares little information to identify the two regions that belong to the same image, even though they have small mutual information. On contrary, the sample pair in the middle image shares too much information to learn a useful representation. Unlike 2D images, there is generally no background in a point cloud, and sample pairs in point clouds have larger CMI. 

CDA divides the pre-training process into $D$ steps and gradually increases the augmentation intensity $\lambda$ and the number of hard samples $N^{c}$ at each step. 
Suppose that a mini-batch
$x$
has $N$ samples, both its $\lambda_{k}$ and $N_{k}^{c}$ at step $k$ are formulated as
\begin{equation}
\lambda _{k}=min \left(
ini \cdot inc^{\left \lfloor \frac{k}{step}\right \rfloor},1 \right ),
\end{equation}
\begin{equation}
N_{k}^{c}= \lambda _{k}\cdot N,
\end{equation}
where $ini$ represents the initial percentage of hard samples, and $inc$ represents an increasing exponential factor used to increase the percentage of the hard samples. $step$ represents the length of iterations in each stage, which can be optionally set fixed. Besides, $k$ denotes the $k$-th step of CDA and $0< k\leq D$. 
Compared to the replicas obtained by only geometric transformation operations, our tactics typically have a higher degree of difficulty for an encoder to identify replicas and original data from the same class.

CDA differs from the traditional use of data augmentation: i) it dynamically influences the latent space to create better embeddings; ii) it enables the model to learn from the hard samples while keeping the difficulty level within the capability of the model during the whole training process.
In general, the more hard samples in the batch, the more difficult that batch is for the model. 
Even if the difficulty differs due to the random cutting of the excised parts (such as the aircraft with the wings removed, which is harder to recognize but the aircraft with the tail removed is relatively easier), the difficulty can be averaged out because of the presence of multiple samples in the batch. 
Since CDA is performed in a self-supervised setting, the increased cost of augmentation can be shared across several tasks instead of just one task in a supervised setting.

\subsection{Curriculum Mutual Information Maximization}

\textbf{Motivation.} We maximize feature-wise and class-wise CMI jointly, rather than a single type of CMI. From a local perspective, the objective of doing joint maximization is to find what is common between two replicas that share redundancy, such as different point clouds of the same object, explicitly encouraging distillation of the common part while ignoring the rest. 
From a global perspective, joint CMI maximization helps to distribute features uniformly in the feature space, gaining strong fault tolerance and improving downstream tasks.

\textbf{Preliminary.}
Given a set $x=	\left\{ x_{1},  \cdots, x_{i},\cdots, x_{N} \right\} $ of $N$ random samples, $x^{a}$ and $x^{b}$ denote random variables from two different replicas augmented by data augmentation $T^{a}$ and $T^{b}$ from CDA.
There are multiple ways to compute mutual information \cite{DBLP:conf/icml/BelghaziBROBHC18, DBLP:conf/iclr/HjelmFLGBTB19}. Herein we estimate CMI by maximizing the variational lower bounds of CMI instead. InfoNCE \cite{DBLP:journals/corr/abs-1807-03748}, as the most common lower bound, is formulated as  
\begin{align}
I(x^{a},x^{b}) & \geq I_{\text{InfoNCE}}(x^{a},x^{b})\label{eq:InfoNCE_1} \\
& \overset{\underset{\mathrm{def}}{}}{=}\mathbb{E}_{p(x_{1:N}^{b})p(x_{i}^{a}|x_{i}^{b})}\left[\log\frac{e^{f(x_{i}^{a},x_{i}^{b})}}{\sum_{\tilde{i}=1}^{N}e^{f(x_{i}^{a},x_{\tilde{i}}^{b})}}\right]+\log N\label{eq:InfoNCE_2}\\
& =\mathbb{E}_{p(x_{1:N}^{b})p(x_{i}^{a}|x_{i}^{b})}\left[\log\frac{e^{S_{ii}}}{\sum_{\tilde{i}=1}^{N}e^{S_{i\tilde{i}}}}\right] +\log N \label{eq:InfoNCE_3} \\
& =-\mathscr{L}_{\text{contrast}}+\log N\label{eq:InfoNCE_4},
\end{align}
where $x_{1:N}^{b}$ are $N$ samples from $p_{x^{b}}$. $x_{i}^{a}$ is a sample from $p_{x^{a}}$ associated with $x_{i}^{b}$ ($i=1,...,N$). $(x_{i}^{a},x_{i}^{b})$
has a strong relationship called positive pair, and $(x_{i}^{a},x_{\tilde{i}}^{b})$ ($\tilde{i}=1,...,N$,$i\neq \tilde{i}$)
are highly independent and called negative pairs. $f(\cdot, \cdot)$ is a similarity function. $S$ is an $N\times N$ similarity matrix where $S_{i\tilde{i}}=f(x_{i}^{a},x_{\tilde{i}}^{a})$. $\mathscr{L}_{\text{contrast}}$ is often known as the contrastive loss \cite{DBLP:conf/icml/ChenK0H20, XinleiChen2020ImprovedBW}.

Since $\log\frac{e^{f(x_{i}^{a},x_{i}^{b})}}{\sum_{\tilde{i}=1}^{N}e^{f(x_{i}^{a},x_{\tilde{i}}^{b})}}\leq0$, the upper bound of $I(x^{a},x^b)$ is $\log N$. We maximize $I(x^{a},x^b)$ to reach the upper bound.
Despite being biased, $I_{\text{InfoNCE}}(x^{a},x^b)$ has much lower variance than other unbiased lower bounds of $I(x^{a},x^b)$ to allow stable model training.

\subsubsection{Feature-wise Curriculum Mutual Information}
\label{sec:3.2.1}
For feature-wise CMI, our goal is to learn an encoder, i.e., a mapping function $E$ that preserves what is common between different replicas and discards instance-specific details. 
Then, each point cloud (i.e., replica) is mapped to a K-dimension assignment feature $f^{a} = E\left ( x^{a} \right )$ and $f^{b} = E\left ( x^{b} \right )$. In terms of network design, our solution is theoretically independent of any particular network. 

As suggested by \cite{DBLP:conf/icml/ChenK0H20}, we do not apply the instance loss to the feature space directly. Instead, we use a projection head $g$ that consists of two-layer MLPs to map the feature to a subspace via $z^{a} = g\left ( f^{a} \right )$ and $z^{b} = g\left ( f^{b} \right )$. The feature representation after $E$ will discard irrelevant information containing data augmentation details, going through the projection head, with the result that most of this extraneous information is filtered out.

$f(\cdot,\cdot)$ can be implemented as the scaled cosine similarity between the representations of $z^{a}$ and $z^{b}$. The similarity matrix is formulated as 
\begin{align}
S^{f}_{ij}=\frac{\left (z_{i}^{a} \right )\left (z_{j}^{b} \right )^{\mathrm{T}}}{\left\|z_{i}^{a} \right\|\left\|z{j}^{b} \right\|}/\tau_{1},
\label{eq:critic_continuous}
\end{align}
where $z_{i}^{a}$ and $z_{j}^{b}$ are the $i$-th and $j$-th rows of $z^{a}$ and $z^{b}$, respectively. $\tau _{1} > 0$
is the feature-wise temperature parameter. 
$\left\|\cdot  \right\|$ is the $\ell_2$-norm, and $\left\Vert z_{i}^{b}\right\Vert =\left\Vert z_{j}^{b}\right\Vert=1$. 

Our goal is to maximize $I(z^{a},z^{b})$.
Thus, we use $f(z_{i}^{a},z_{j}^{b})$
instead of $f(x_{i}^{a},x_{\tilde{i}}^{b})$ in Eq.~\ref{eq:critic_continuous} to emphasize that it is a function of representations in this context.
Regarding this, we rewrite the contrastive loss in Eq.~\ref{eq:InfoNCE_4} as
\begin{equation}
\begin{split}
\mathscr{L}_{Fea}^{a} 
& =\mathbb{E}_{p(z_{1:N}^{b})p(z_{i}^{a}|z_{i}^{b})}\left[f(z_{i}^{a},z_{i}^{b})-\log\sum_{j=1}^{N}\exp(f(z_{i}^{a},z_{j}^{b}))\right]\label{eq:contrast_continuous_2} \\
& =\mathbb{E}_{p(z_{1:N}^{b})p(z_{i}^{a}|z_{i}^{b})}\left[\log\frac{e^{S^{f}_{ii}}}{\sum_{j=1}^{N}e^{S^{f}_{ij}}}\right],
\end{split}
\end{equation}
where $Fea$ stands for the feature.
Similarly, we can obtain $\mathscr{L}_{Fea}^{b}$ for any $z_{i}^{b}$.
The feature loss can be formulated as
\begin{equation}
    \mathscr{L}_{Fea}=\frac{1}{2N}\left ( \mathscr{L}_{Fea}^{a}+\mathscr{L}_{Fea}^{b} \right ).
\end{equation}

\subsubsection{Class-wise Curriculum Mutual Information}
For the perspective of class, the two batches $x^{a}$ and $x^{b}$ should have the same sample distribution. Therefore, samples classified as the same class are viewed as positives. Similar to the projection head $g$, we use another two-layer MLP followed by softmax to form the class head $c$. Then, we project the feature into an $M$-dimensional space, where $M$ is equal to or bigger than the number of classes in the pre-trained dataset. We obtain features $y^{a}$ and $y^{b}$ via $y^{a}=c\left ( f^{a} \right )$ and $y^{b}=c\left ( f^{b} \right )$. The output $\left [ 0,1 \right ]^{M}$ is interpreted as the distribution of a discrete random variable $y$ over $M$ classes. $y^{a}_{i}$ is the $i$-th column of $Y^{a}$, i.e., the representation of class $i$ under the first data augmentation. The second augmented representation of class $i$ is $y^{b}_{i}$ (the $i$-th column of $y^{b}$). 
Not only $x^{a}_{i}$ and $x^{b}_{i}$ should belong to the same class, but also $y^{a}_{i}$ and $y^{b}_{i}$ should have the same class distribution. For $y^{a}_{i}$, the class loss is
\begin{equation}
\begin{split}
\mathscr{L}_{Cls}^{a} &
=\mathbb{E}_{p(y_{1:M}^{b})p(y_{i}^{a}|y_{i}^{b})}\left[\log\frac{e^{S^{c}_{ii}}}{\sum_{j=1}^{M}e^{S^{c}_{ij}}}\right],
\label{eq:contrast_continuous}
\end{split}
\end{equation}
which has one positive class and $M-1$ negative classes.

Similar to Eq.~\ref{eq:critic_continuous}, the similarity matrix of $(y_{a},y_{b})$ is given by
\begin{equation}
S_{ij}^{c} =\frac{\left (y_{i}^{a} \right )\left (y_{j}^{b} \right )^{\mathrm{T}}}{\left\|y_{i}^{a} \right\|\left\|y_{j}^{b} \right\|}/\tau_{2},
\label{eq:contrast_continuous_3}
\end{equation}
where $y_{i}^{a}$ and $y_{i}^{a}$ are the $i$-th and $j$-th columns of $z^{a}$ and $z^{b}$, respectively. $\tau _{2}$ 
is the class-wise temperature parameter. After going over every class, the class loss is determined as
\begin{equation}
    \mathscr{L}_{Cls}=\frac{1}{2M}\left ( \mathscr{L}_{Cls}^{a}+\mathscr{L}_{Cls}^{b} \right ).
\end{equation}
\subsubsection{Total Loss}
The overall objective function of PointSmile is defined as
\begin{equation}
\mathscr{L}_{overall}=\mathscr{L}_{Fea}+\mathscr{L}_{Cls}+ \beta \mathscr{L}_{CR},
\end{equation}
where $\mathscr{L}_{CR}$ is a class regularization loss \cite{LukasMeier2008TheGL} defined as 
\begin{equation}
 \mathscr{L}_{CR}=\frac{1}{N}\sum_{M}^{i=0}\left (\sum_{N}^{j=0}y_{i}^{j}  \right ).
\end{equation}
$\mathscr{L}_{CR}$ can avoid local optimum solutions or degenerated solutions where all samples fall into the same class (e.g., $y$ is one-hot for all samples). $\beta$ is a coefficient.


\section{Experiments}

\label{sec:experiments}
To learn features transferred effectively to downstream tasks is the primary goal of representation learning. We employ three downstream tasks including classification, part segmentation, and semantic segmentation, to assess the transferability of PointSmile and its competitors. We follow the same procedure as \cite{DBLP:conf/icml/AchlioptasDMG18}, and we train our model on ShapeNet \cite{DBLP:journals/corr/ChangFGHHLSSSSX15}. 
Ablation studies are also presented to analyze the effectiveness of both curriculum data augmentation and joint curriculum mutual information.

\begin{figure*}[h]
    \centering
    \includegraphics[width=1\textwidth]{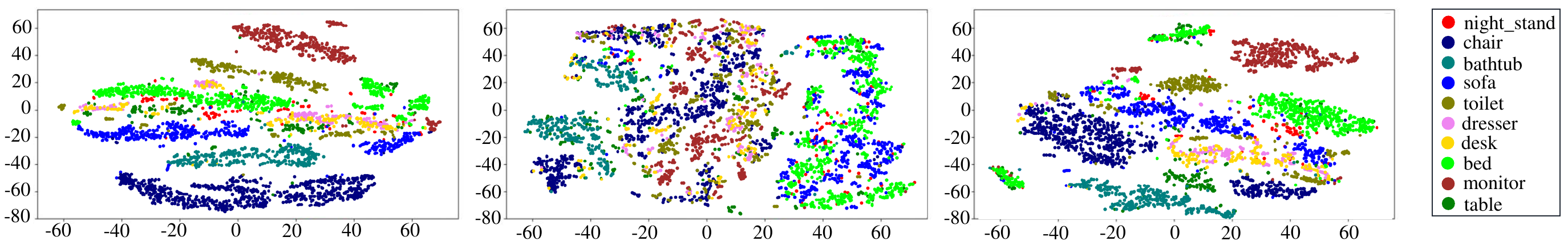}
    \caption{t-SNE visualization of the features learned from ModelNet10 after training PointNet as the backbone. The features learned by maximizing both CMI (right) provide better discrimination of classes than using only feature-wise CMI (left) or class-wise CMI (middle).
     }
    \label{fig:tsne}
\end{figure*}

\begin{table*}
\fontsize{9}{9}\selectfont
\centering

\caption{Comparison of the linear SVM classification on ModelNet40 \cite{DBLP:conf/cvpr/WuSKYZTX15}. `Self-sup.', `Decoder-free.', `Sup.', and `Fine.' denote the models pre-trained in a self-supervised, without reconstruction module, supervised, and finetuned manner, respectively.}

\setlength{\tabcolsep}{5.7mm}{
\renewcommand{\arraystretch}{1.5}
\begin{tabular*}{\linewidth}{cccc|cc}
\toprule
Self-sup.   & Decoder-free                         & \multicolumn{1}{c}{PointNet Acc.} & \multicolumn{1}{c}{DGCNN Acc.} & \multicolumn{1}{c}{Sup.}                & \multicolumn{1}{c}{Acc.}          \\ \toprule
DeepCluster \cite{DBLP:conf/eccv/CaronBJD18}   &    \checkmark                                      & 86.3                              & 90.4                            & PointNet\cite{DBLP:conf/cvpr/QiSMG17}                                  & 89.2          \\
Jigsaw3D \cite{DBLP:conf/nips/SauderS19}                                       &  $\times$    & 87.3                              & 90.6                            & PointNet++ \cite{DBLP:conf/nips/QiYSG17}                            & 90.7          \\
Rotation3D  \cite{DBLP:conf/3dim/PoursaeedJQXK20}                                  &   \checkmark     & 88.9                              & 90.8                            & DGCNN \cite{DBLP:journals/tog/WangSLSBS19}                                  & 92.9          \\
Info3D \cite{DBLP:conf/eccv/Sanghi20}        &   \checkmark                 & 89.8                              & 91.6                            & PointTransformer \cite{DBLP:conf/iccv/ZhaoJJTK21}                       & 93.7          \\ 
\cline{5-6} 
OcCo \cite{DBLP:conf/iccv/WangLYLK21}                                     &   \checkmark      & 88.7                              & 90.2                            & Fine.                                & Acc.          \\ \cline{5-6} 
STRL \cite{DBLP:conf/iccv/HuangXZZ21}            &   \checkmark   & 88.3                              & 90.0                            & Transformer-OcCo \cite{DBLP:conf/iccv/WangLYLK21}                  & 92.1          \\
ParAE \cite{DBLP:conf/cvpr/EckartYLK21}      &        $\times$                               & \textbf{90.3}                     & 91.6                            & Point-Bert\cite{DBLP:conf/cvpr/YuTR00L22}                     & \textbf{93.2} \\
CrossPoint \cite{DBLP:conf/cvpr/AfhamDDDTR22}         & \checkmark                                  & 89.1                              & 91.2                            & PointSmile (PointNet)                        & 90.7          \\
\multicolumn{1}{l}{PointSmile (ours)}      & \checkmark                 & 90.0                    & \textbf{91.8}                   & PointSmile (DGCNN)                         & 93.0          \\ \bottomrule
\end{tabular*}}
\label{tab:classificationM}
\end{table*}

\begin{table}[t]
	\begin{minipage}{.50\textwidth}
		\centering
		\caption{Comparison of classification on ScanObjectNN \cite{DBLP:conf/iccv/UyPHNY19}. }
		
  \large
  \renewcommand{\arraystretch}{1.1}
			\begin{tabular}{ccc}
        
			\toprule
			    Encoder & Method & Acc. \\
				\midrule
				\multirow{4}{*}{PointNet}
				& Sup.\cite{DBLP:conf/cvpr/QiSMG17} & 83.7\\
				& OcCo \cite{DBLP:conf/iccv/WangLYLK21} & 69.5\\
				& CrossPoint \cite{DBLP:conf/cvpr/AfhamDDDTR22} & 75.6\\
				& PointSmile (ours) & \textbf{75.8}\\
				\midrule
				\multirow{4}{*}{DGCNN}
				& Sup.\cite{DBLP:journals/tog/WangSLSBS19} & 85.2\\
				& OcCo \cite{DBLP:conf/iccv/WangLYLK21} & 69.5\\
			     & CrossPoint \cite{DBLP:conf/cvpr/AfhamDDDTR22} & 81.7\\
				& PointSmile (ours) & \textbf{82.8}\\
				\bottomrule
			\end{tabular}
\label{tab:classificationS}
\end{minipage}
\end{table}

\begin{figure*}[]
    \centering
    \includegraphics[width=1\textwidth]{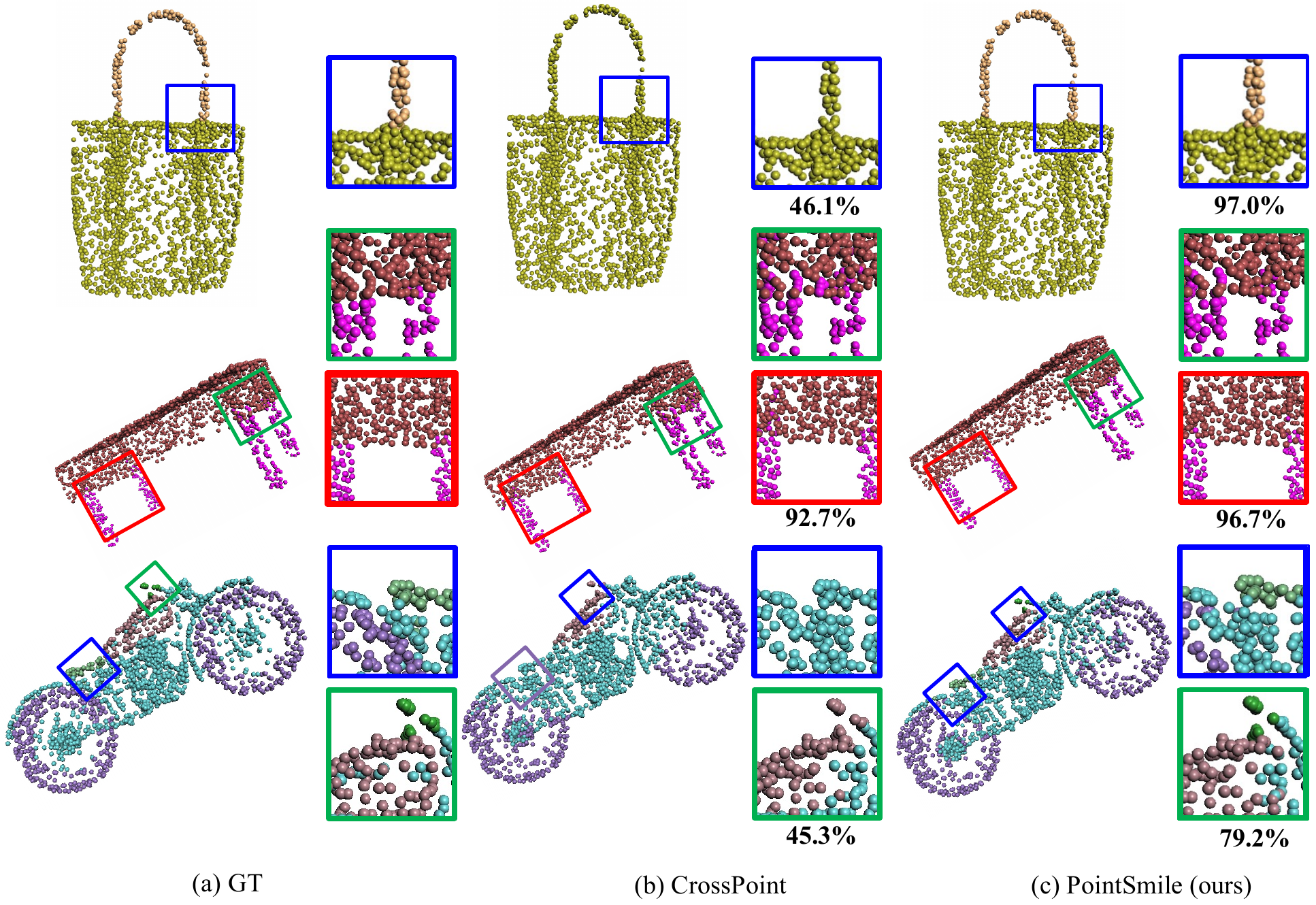}
    \caption{Segmentation results on ShapePart of CrossPoint \cite{DBLP:conf/cvpr/AfhamDDDTR22} and PointSmile (DGCNN as the encoder). Different colors represent different parts. }
    \label{fig:seg}
\end{figure*}

\begin{figure*}[]
    \centering
    \includegraphics[width=1\textwidth]{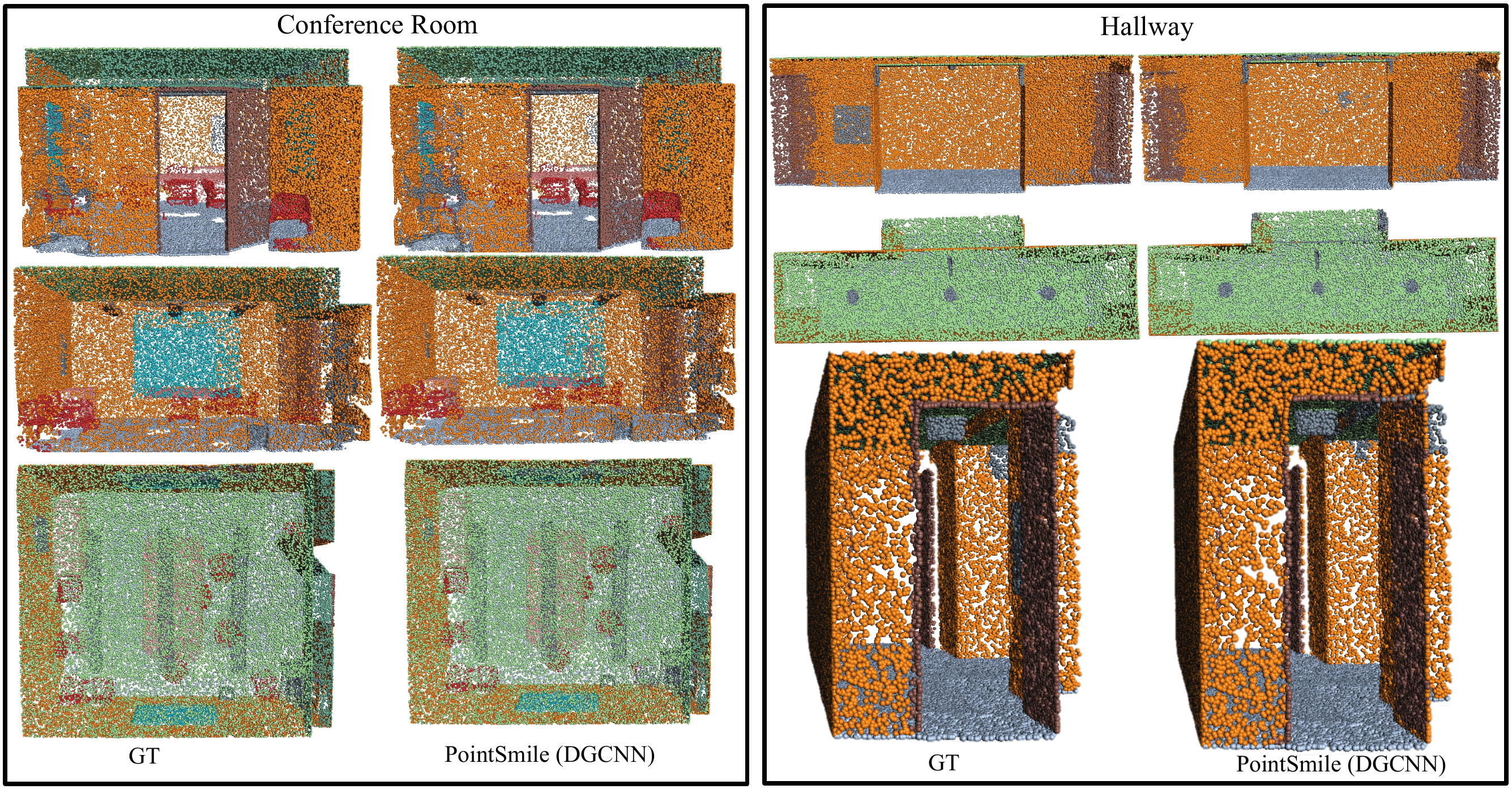}
    \caption{Semantic segmentation results (DGCNN as the backbone) on S3DIS. Different colors indicate different objects. We show two different indoor scenes in two boxes. For each scene we show three views (front, side and back).}
    \label{fig:semseg}
\end{figure*}

\subsection{Pre-training Setup}
To fairly compare our PointSmile with existing techniques, we adopt PointNet \cite{DBLP:conf/cvpr/QiSMG17} and DGCNN \cite{DBLP:journals/tog/WangSLSBS19} as our feature extractors. 
For the projection head, we employ a 2-layer MLP to yield a 256-dimensional feature vector projection in the invariant space, and the cluster head is as well. 
For using PointNet as the backbone, we employ the SGD optimizer \cite{JMichaelCherry1998SGDSG} with an initial learning rate of $1\times 10^{-3}$, the momentum of 0.9 and the weight decay of $1\times 10^{-6}$. The learning rate scheduler is cosine annealing \cite{IlyaLoshchilov2016SGDRSG}, and the model is trained end-to-end across 200 epochs. The training takes about 33 hours on a RTX 3060 GPU. For using DGCNN as the backbone, we use the ADAM  optimizer \cite{DiederikPKingma2014AdamAM} with the weight decay of $1\times 10^{-4}$ and the initial learning rate of $1\times 10^{-3}$. It takes roughly 124 hours to complete the training on RTX 3090 GPU. 

\subsection{Downstream Task Setup}

\textbf{Object classification.}
Given an object represented by a set of points, object classification predicts the class to which the object belongs. We use two benchmarks: ModelNet40 \cite{DBLP:conf/eccv/SharmaGF16} and ScanObjectNN \cite{DBLP:conf/iccv/UyPHNY19}. We perform our synthetic object classification experiments on ModelNet40. ModelNet40 is composed of 12331 meshed models from 40 object categories, split into 9843 training meshes and 2468 testing meshes, on which the points are sampled. ScanObjectNN is a demanding and realistic 3D point cloud classification benchmark dataset made up of occluded objects captured from real indoor scenes which is more challenging. It contains 2,880 objects (2304 for training and 576 for testing) from 15 categories. We use the same settings as \cite{DBLP:conf/cvpr/QiSMG17, DBLP:journals/tog/WangSLSBS19} for fine-tuning. 

Specifically, for PointNet, we use the Adam optimizer with an initial learning rate of 1e-3, and the learning rate is decayed by 0.7 every 20 epochs with the minimum value of 1e-5. For DGCNN, we use the SGD  optimizer \cite{JMichaelCherry1998SGDSG} with the momentum of 0.9 and the weight decay of 1e-4. The learning rate starts from 0.1 and then decays using cosine annealing \cite{IlyaLoshchilov2016SGDRSG} with the minimum value of 1e-3. We use dropout \cite{srivastava2014dropout} in the fully connected layers before the softmax output layer. The dropout rate is set to 0.7 for PointNet and is set to 0.5 for DGCNN. For all the models, we train them for 200 epochs with a batch size of 32.

\textbf{Part segmentation.}
Part segmentation is a challenging fine-grained 3D recognition task. The mission is to predict the part category label (e.g., car wheel, bag handle) of each point for a given object. For 3D object part segmentation, we choose ShapeNetPart \cite{DBLP:journals/tog/YiKCSYSLHSG16} that contains 16881 objects of 2048 points from 16 categories with 50 parts in total. Following PointNet \cite{DBLP:conf/cvpr/QiSMG17}, we sample 2,048 points from each model. For PointNet, we use the Adam optimizer with an initial learning rate of 1e-3, and the learning rate is decayed by 0.5 every 20 epochs with the minimum value of 1e-5. For DGCNN, we use an SGD optimizer with the momentum of 0.9 and the weight decay of 1e-4. The learning rate starts from 0.1 and then decays using cosine annealing with the minimum value of 1e-3. We train the models for 250 epochs with a batch size of 16.

\textbf{Semantic segmentation.}
Semantic segmentation predicts the semantic object category of each point. We evaluate PointSmile on semantic segmentation on Stanford Large Scale 3D Indoor Spaces (S3DIS) \cite{armeni20163d}. S3DIS consists of 3D scans collected by Matterport scanners from 6 indoor areas, containing 271 rooms and 13 semantic classes. We train all models for 100 epochs with a batch size of 24.
\subsection{Evaluation on Downstream Tasks}

\begin{table}[t]
	\begin{minipage}{.49\textwidth}
		\centering
		\caption{Comparison of the overall accuracy and mean IoU of part segmentation results with self-supervised methods on ShapeNetPart \cite{DBLP:journals/tog/YiKCSYSLHSG16}.}
		\resizebox{.90\textwidth}{!}{
            \renewcommand{\arraystretch}{1.1}
			\begin{tabular}{clcccc}
				\toprule
				\multirow{2}{*}{Encoder} & 
				\multirow{2}{*}{Method} &
				\multicolumn{2}{c}{Metrics} \\
				\cline{3-4} 
				\multicolumn{2}{c}{} & OA ($\%$) & mIoU ($\%$) \\
				\toprule
				\multirow{4}{*}{PointNet} 
				& Jigsaw3D \cite{DBLP:conf/nips/SauderS19} & 93.1 & 82.2 \\
				& OcCo \cite{DBLP:conf/iccv/WangLYLK21} & 93.4 & 83.4 \\
				& CrossPoint \cite{DBLP:conf/cvpr/AfhamDDDTR22}  & 93.2 & 82.7 \\
				& PointSmile (ours) & \textbf{93.6} & \textbf{83.5} \\
				\midrule
				\multirow{4}{*}{DGCNN} 
				& Jigsaw3D \cite{DBLP:conf/nips/SauderS19} & 92.7 & 84.3 \\
				& OcCo \cite{DBLP:conf/iccv/WangLYLK21}  & 94.4 & 85.0 \\
				& CrossPoint \cite{DBLP:conf/cvpr/AfhamDDDTR22}  & 94.4 & 85.3 \\
				& PointSmile (ours) & \textbf{94.4} & \textbf{85.4} \\
				\bottomrule
			\end{tabular}
		}
        \label{tab:part_segmentation}
	\end{minipage}
\end{table}

\begin{table}[t]
	\begin{minipage}{.49\textwidth}
		\centering
		\caption{Comparison of the overall accuracy and mean IoU of semantic segmentation results with self-supervised methods on  S3DIS \cite{DBLP:journals/tog/YiKCSYSLHSG16}.}
		\resizebox{.90\textwidth}{!}{
			\begin{tabular}{clcccc}
				\toprule
				\multirow{2}{*}{Encoder} & 
				\multirow{2}{*}{Method} &
				\multicolumn{2}{c}{Metrics} \\
				\cline{3-4} 
				\multicolumn{2}{c}{} & OA ($\%$) & mIoU ($\%$) \\
				\toprule
				\multirow{5}{*}{PointNet} 
				& Jigsaw3D \cite{DBLP:conf/nips/SauderS19} & 80.1 & 52.6 \\
				& OcCo \cite{DBLP:conf/iccv/WangLYLK21} & 82.0 & 54.9 \\
				& CrossPoint \cite{DBLP:conf/cvpr/AfhamDDDTR22}  & 81.8 & 54.5 \\
				& PointSmile (ours) & \textbf{82.4} & \textbf{55.0} \\
				\midrule
				\multirow{5}{*}{DGCNN} 
				& Jigsaw3D \cite{DBLP:conf/nips/SauderS19} & 84.1 & 55.6 \\
				& OcCo \cite{DBLP:conf/iccv/WangLYLK21}  & 84.6 & 58.0 \\
				& CrossPoint \cite{DBLP:conf/cvpr/AfhamDDDTR22}  & 87.4 & 58.4 \\
				& PointSmile (ours) & \textbf{86.9} & \textbf{58.9} \\
				\bottomrule
			\end{tabular}
		}
        \label{tab:sem_segmentation}
	\end{minipage}
\end{table}

\begin{figure}[t]
    \centering
    \includegraphics[width=0.47\textwidth]{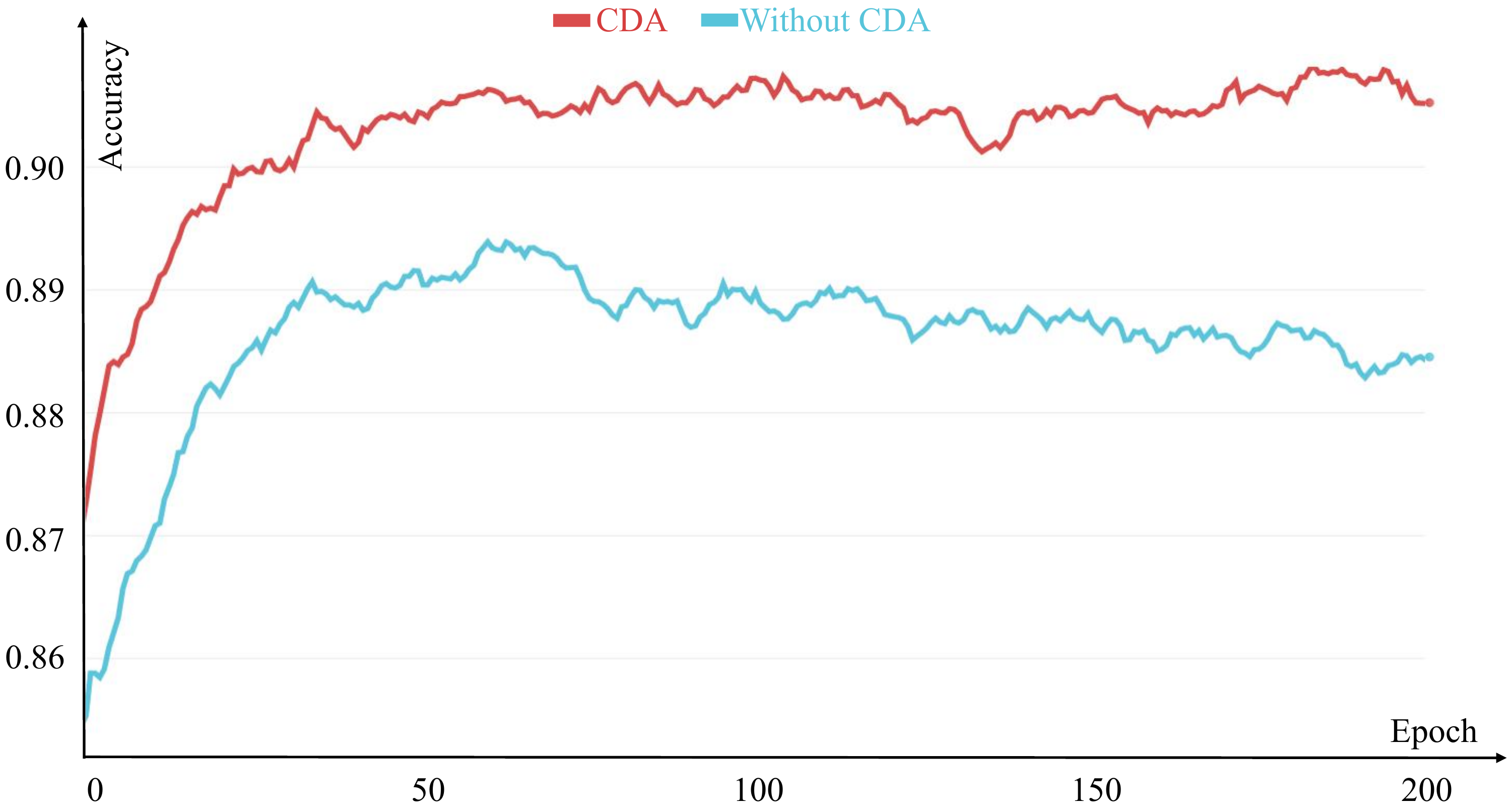}
    \caption{The effect of curriculum data augmentation (CDA) on classification of the models in ModelNet40 \cite{DBLP:conf/eccv/SharmaGF16} using PointNet \cite{DBLP:conf/cvpr/QiSMG17} as the backbone.}
    \label{fig:losses}
\end{figure}

\begin{table}[]
	\begin{minipage}{.47\textwidth}
		\centering
		\small
		\caption{The accuracy of linear SVM classification using retrained embedding on  ModelNet40 \cite{DBLP:conf/eccv/SharmaGF16} for PointSmile. }
		\resizebox{.80\textwidth}{!}{
			\begin{tabular}{cccc}
			\toprule
			    Encoder & $L_{Fea}$ & $L_{Cls}$ & Acc. \\
				\midrule
				\multirow{3}{*}{PointNet}
				& \checkmark & & 88.2\\
				&  &\checkmark& 87.6\\
				& \checkmark & \checkmark & \textbf{90.0}\\
				\midrule
				\multirow{3}{*}{DGCNN}
				& \checkmark & & 91.1\\
				& &\checkmark& 90.7\\
				& \checkmark & \checkmark & \textbf{91.8}\\
				\bottomrule
			\end{tabular}}
\label{tab:nagetive compare}
\end{minipage}
\end{table}


\subsubsection{Synthetic Object Classification}
After equally sampling each object with 1,024 points, the  coordinates $\left ( x,y,z \right )$ of sampled points are used as input for the classification task. Following each of the methods in the standard experimental process outline \cite{DBLP:conf/iccv/WangLYLK21,DBLP:conf/nips/SauderS19}, we train a simple linear SVM (Support Vector Machine) classifier \cite{DBLP:journals/ml/CortesV95} using the extracted 3D point cloud features, while disabling the pre-trained point cloud feature extractor, to assess the utility of the feature representations in classification. 

As shown in Table~\ref{tab:classificationM}, our method outperforms the state-of-the-art methods on ModelNet40, no matter PointNet or DGCNN is employed.
Please note that CrossPoint \cite{DBLP:conf/cvpr/AfhamDDDTR22} necessitates multi-modal data support, while our PointSmile is single-modal based. Still, PointSmile outperforms CrossPoint by a margin of 0.9\% and 0.6\% by using the backbone of PointNet and DGCNN, respectively. 
Furthermore, our PointSmile using a simple backbone like PointNet, prevails over many self-supervised methods with complex architectures, and it also surprisingly demonstrates superior performance than the original PointNet supervised learning benchmark. 
It is also shown that initializing our model with pre-trained weights also helps in achieving high accuracy in a fine-tuned manner. Even without a Transformer framework, it can achieve a nearly flat accuracy with Point-Bert.

\subsubsection{Real-world Object Classification}
To validate the effectiveness of our method on real-world point clouds, we perform classification experiments on ScanObjectNN \cite{DBLP:conf/iccv/UyPHNY19}.
We use a simple linear SVM for classification. Table~\ref{tab:classificationS} reports the linear evaluation results on ScanObjectNN. Compared with the state-of-the-art self-supervised methods, the accuracy for the DGCNN backbone is significantly improved by 1.6\%, indicating that the feature representation learned by our PointSmile can span from synthetic data to realistic real-world settings.

\subsubsection{Part Segmentation}
The part classifier is built by using the same architecture as those used in PointNet and DGCNN for part segmentation. Our self-supervised model is initially trained on ShapeNet and then fixed as a feature extractor. The evaluation metrics are OA (Overall Accuracy) and mIoU (mean Intersection over Union).

Table~\ref{tab:part_segmentation} summarizes the evaluation results and the comparison of our PointSmile against several alternative methods on ShapeNetPart. Our method improves the part segmentation performance and exceeds the state-of-the-art baselines in terms of OA and mIoU. Furthermore, compared to other self-supervised approaches, our model performs better on the backbone of PointNet and DGCNN, demonstrating that more discriminative features can be learned by maximizing feature- and class-wise CMI. 

Moreover, we observe from the results that our model can capture essential fine-grained features. Figure~\ref{fig:seg} demonstrates some of the qualitative part segmentation results  showing that our method can achieve an excellent performance for part segmentation. We find that our method is  able to better segment fine details than other methods in various categories. For example, the handle of bag, the tail of airplane and the wheel of motorcycle  can be all segmented clearly from other parts. 

\subsubsection{Semantic Segmentation}
Semantic segmentation is a technique for associating points or voxels with semantic object labels, and it is also a fundamental research challenge in point cloud processing. 
As shown in Table~\ref{tab:sem_segmentation}, with the PointNet as an encoder, PointSmile achieves 82.9\% OA and 55.3\% mIoU, outperforming the excellent baselines in terms of OA and mIoU. With the DGCNN as an encoder, PointSmile achieves 85.4\% OA and 59.2\% mIoU, outperforming CrossPoint, OcCo and Jigsaw3D.
Figure~\ref{fig:semseg} shows the visualization results, where our results are very close to the ground-truths in the different scenes. Besides, PointSmile performs well when segmenting small objects in the scene, such as the clutter on the ceiling.


\subsection{Ablations and Analysis}

\subsubsection{Curriculum Data Augmentation}

Existing research in image processing has demonstrated that a powerful data augmentation technique is crucial for downstream tasks \cite{DBLP:conf/icml/ChenK0H20}. We also discover that a more sophisticated backbone requires more challenging beginning samples. To verify the significance of curriculum data augmentation (denoted as CDA), we test our model on ModelNet40 by omitting CDA using the backbone of PointNet \cite{DBLP:conf/cvpr/QiSMG17}. Figure~\ref{fig:losses} shows that CDA enhances the performance of PoineSmile. In addition, with CDA, the accuracy can remain stable and consistently rise without decreasing at a later stage, according to the resulting curve.

\subsubsection{Joint Curriculum Mutual Information}
To examine the effectiveness of our design, we perform ablation study on ModelNet40 by using (i) only the feature-wise CMI, (ii) only the class-wise CMI, and (iii) both CMI, to understand their efficacy. The results are reported in Table~\ref{tab:nagetive compare}. The feature-wise CMI model gets a classification accuracy of 91.1\% for DGCNN and 88.2\% for PointNet. Maximizing class-wise CMI can
significantly improve the accuracy of the baseline model, which increases nearly by 0.8\% and 0.7\% for PointNet and DGCNN, respectively. However, utilizing only the class-wise CMI achieves the lowest performance, due to the mis-allocation of samples that belong to the same category but do not have similar features. Using both types of CMI produces a better outcome than using only one of them. This test reveals that the two types of CMI interact with and complement each other. 

The t-SNE plot of the features trained by PointNet as the backbone on ModelNet10 \cite{DBLP:conf/eccv/SharmaGF16} 
is shown in Figure~\ref{fig:tsne}. The left part of Figure~\ref{fig:tsne} is only with feature-wise CMI, showing that points of the same color are almost on the same level, but points of different colors are also farther apart. The middle part is only with class-wise CMI, showing that the points of different colors are more evenly distributed, but there are no clear boundaries. This is detrimental to downstream tasks. The right part is the joint CMI, showing that points of the same color are clustered together and uniformly distributed over the feature space. 

It is clear that even without labeled data, both feature- and class-wise settings yield some degrees of class discrimination. Besides, combining the two improves the discrimination boundaries in these classes, resulting in superior discriminative features.

\section{Conclusion}
We propose a novel self-supervised learning paradigm, namely PointSmile, for representation learning on point clouds. PointSmile is well designed to mimic how humans to learn new yet difficult knowledge in a ``how-and-what-learn" manner. To this end, we design a curriculum in which the `learner' studies from easy samples to hard samples gradually. In the study, PointSmile can exploit both the fine-grained feature-level and coarse-grained class-level information via maximizing curriculum mutual information. 
PointSmile extracts discriminative features of point clouds better than its competitors, which is verified by downstream tasks including object classification and segmentation. 


%





\ifCLASSOPTIONcaptionsoff
  \newpage
\fi



\bibliographystyle{IEEEtran}
\bibliography{egbib}
\end{document}